# EVALUATION OF DEEP LEARNING SEMANTIC SEGMENTATION FOR LAND COVER MAPPING ON MULTISPECTRAL, HYPERSPECTRAL AND HIGH SPATIAL AERIAL IMAGERY

Ilham Adi Panuntun[1], Ying-Nong Chen*[2], Ilham Jamaluddin[3] and Thi Linh Chi Tran[1]

[1]Graduate Student, Center for Space and Remote Sensing Research, National Central University,
No. 300, Jhongda Rd., Jhongli Dist., Taoyuan City 320317, Taiwan
Email: ilhamadipanuntun@g.ncu.edu.tw; ttlchi@g.ncu.edu.tw

[2]Assistant Professor, Center for Space and Remote Sensing Research, National Central University,
No. 300, Jhongda Rd., Jhongli Dist., Taoyuan City 320317, Taiwan
Email: yingnong1218@csrsr.ncu.edu.tw

[3]Graduate Student, Department of Computer Science and Information Engineering, National Central University,
No. 300, Jhongda Rd., Jhongli Dist., Taoyuan City 320317, Taiwan
Email: ilhamjamaluddin@g.ncu.edu.tw

**ABSTRACT:** In the rise of climate change, land cover mapping has become such an urgent need in environmental monitoring. The accuracy of land cover classification has gotten increasingly based on the improvement of remote sensing data. Land cover classification using satellite imageries has been explored and become more prevalent in recent years, but the methodologies remain some drawbacks of subjective and time-consuming. Some deep learning techniques have been utilized to overcome these limitations. However, most studies implemented just one image type to evaluate algorithms for land cover mapping. Therefore, our study conducted deep learning semantic segmentation in multispectral, hyperspectral, and high spatial aerial image datasets for landcover mapping. This research implemented a semantic segmentation method such as Unet, Linknet, FPN, and PSPnet for categorizing vegetation, water, and others (i.e., soil and impervious surface). The LinkNet model obtained high accuracy in IoU (Intersection Over Union) at 0.92 in all datasets, which is comparable with other mentioned techniques. In evaluation with different image types, the multispectral images showed higher performance with the IoU, and F1-score are 0.993 and 0.997, respectively. Our outcome highlighted the efficiency and broad applicability of LinkNet and multispectral image on land cover classification. This research contributes to establishing an approach on landcover segmentation via open source for long-term future application.

## 1. Introduction

In the rise of climate change, land cover mapping has become such an urgent need in environmental monitoring. The significance of precise and punctual land cover maps cannot be overstated, as they serve many purposes in various fields, including urban and regional planning, monitoring disasters and hazards, managing natural resources and the environment, and ensuring food security (Zhang et al., 2022). The utilization of land cover mapping has the potential to address various substantial concerns on a broad scale, including but not limited to global warming, the rapid decline of species' habitats, unprecedented population migration, the escalating process of urbanisation, and the widening disparities within and among nations (Hasan et al., 2020). However, due to the change of land use, environmental variations, and human activities, the high accuracy in land cover mapping has been still a challenging problem.

The accuracy of land cover classification has become increasingly based on the essential of remote sensing data. Nevertheless, the methodologies remain some drawbacks of subjective and time-consuming (Tzepkenlis et al., 2023; Shekar et al., 2023). Meanwhile, this method has still been widely implemented in recent studies (Hussain et al., 2023; Seyam et al., 2023; Wahla et al., 2023). Hence, efficiency of land cover mapping via remote sensing approach still remains weaknesses.

Deep learning has been conducted and recommended as the approach to overcome these limitations. In recent decades, analyzing remote sensing imagery by deep learning (DL) has become a new trend of methodology in the field of environmental monitoring (Xue et al., 2023; Dimyati et al., 2023; Tran et al., 2022; Jamaluddin et al., 2021; Chen et al, 2020). As a significant offshoot of machine learning characterized by utilizing several layers for processing, it can potentially enhance model performance and accuracy (LeCun et al., 2015). There exist three distinct types of deep learning (DL) algorithms that are commonly conducted for the evaluation of satellite data. These include the patch-based convolutional neural network (CNN), semantic segmentation, also known as the Fully Convolutional Network (FCN), and object detection methods (Neupane et al., 2021; Hoeser et al., 2020).

Research approaches via semantic segmentation have taken many scientists' attention and expanded in recent studies. Those publications have utilized various networks for different tasks. These networks include U-Net (Ronneberger et al., 2015; Al-Saad et al., 2023; Wang et al., 2023), SegNet (Yang et al., 2020; Torres et al., 2020), DeepLabV3+ (Girisha et al., 2019; Torres et al., 2020), Adversarial Networks (Li et al., 2019), FCN-ConvLSTM (Wang et al., 2019), Linknet (Chaurasia et al., 2017; Ai et al., 2023), FPN (Lin et al., 2017; Seferbekov et al., 2018; Li et al., 2023), PSPNet (Zhao et al., 2017; Sun et al., 2023) and more. Particularly, the deep learning semantic segmentation in land cover classification showed up a high performance in using multispectral and multi-temporal data (Du et al., 2021). Thus, semantic segmentation shows its potential to enhance prior methods in the field of land cover mapping.

The primary objective of this research is to overcome the limitations of conventional methods and identify the most optimal semantic segmentation architecture for improved land cover mapping using various remote sensing imagery. Our study implemented different semantic segmentation methods such as Unet, Linknet, FPN, and PSPnet for categorizing vegetation, water, and others (i.e., soil and impervious surface). We further analyzed different remote sensing imagery consisting of multi-spectral images, hyper-spectral images and high spatial images. This research contributes to establishing an approach on land cover segmentation via open source for long-term future application.

## 2. Method

### 2.1 Study area and dataset

There are 3 study sites: 1) Taipei region consist of 12 bands of Landsat 8 data with spatial resolution of 30m, 2) Pavia University, and 3) Hamlin Beach State Park (Figure 1). The Taipei site used Landsat 8 satellite data consisting of 8 original Landsat 8 bands (Aerosol, Blue, Green, Red, Near Infrared (NIR), Shortwave Infrared 1 (SWIR1), Shortwave Infrared 2 (SWIR2), and Thermal Infrared 1 (TIRS1)) and 4 spectral indices (Normalized Difference Water Index (NDWI), Normalized Different Vegetation Index (NDVI), Normalized Different Bare Soil Index (NDBSI), and Normalized Difference Built-up Index (NDBI)) at 30 m resolution. The Pavia University site is hyperspectral imagery and used ROSIS satellite data that consists of 103 bands at 1.3 m resolution. The Hamlin Beach State Park site is aerial imagery and used UAV data consists of 6 bands (Red, Green, Blue, NIR1, NIR2, NIR3) at 0.047 m resolution. The images of Taipei, Pavia University and Hamlin Beach State Park was first cropped into small blocks size as 64 x 64, 32 x 32 and 64 x 64 respectively. Dataset of each region was separated as 70% for training, 15% for validation and 15% for testing. The target data used in this study were obtained from three different sources. The multispectral data was processed using the LSMA-PPM method as described by (Zhao et al., 2022). The hyperspectral data was generated using the SpectralFormer, as established by (Hong et al., 2021). Lastly, the high spatial aerial imagery was obtained from the RIT-18 dataset, as documented by (Kemker et al., 2018).

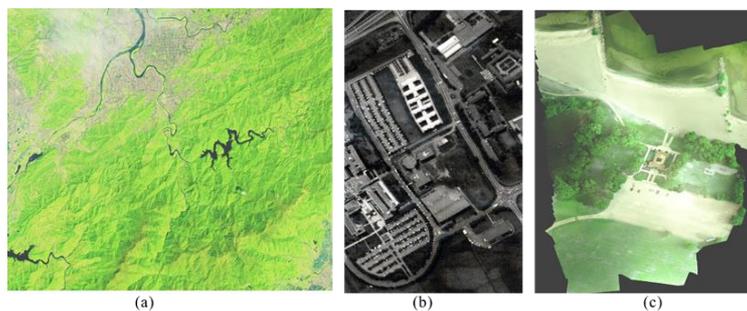

Figure 1. Aerial image of study regions, (a) Multispectral images (Landsat 8 satellite) of Taipei, (b) Hyperspectral image (ROSIS satellite) of Pavia University, and (c) Multispectral image (UAV data) of Hamlin Beach State Park.

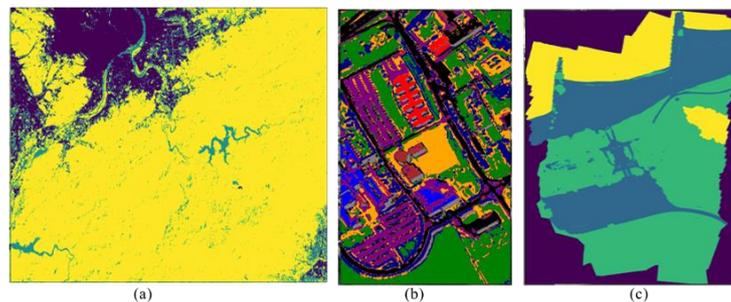

Figure 2. Ground truth, (a) Multispectral images (Landsat 8 satellite) of Taipei, (b) Hyperspectral image (ROSIS satellite) of Pavia University, and (c) Multispectral image (UAV data) of Hamlin Beach State Park.

## 2.2 Semantic Segmentation Architectures:

In our study to evaluate the segmentation of Land Cover, we employed four advanced neural network models: Unet (Ronneberger et al., 2015), Linknet (Chaurasia et al., 2017), FPN (Lin et al., 2017), and PSPNet (Zhao et al., 2017). The network architectures are presented in Figure 3. The networks can be categorized as decoder-coder models with skip connections based on convolutional neural networks (CNNs), operating at many scales. Encoding involves extracting features from the incoming data while reducing the width and height dimensions. In contrast, the decoder component increases the resolution of feature maps while reducing the number of feature channels by half. Skip connections have been found to mitigate the issue of vanishing gradients (He et al., 2016), enhancing the learning process's efficacy. In conclusion, the output layer employs a 1x1 convolution operation with an activation function to transform the feature vector into an output matrix with the same width and height as the input image. It contains a corresponding number of channels as distinct classes. The activation function employed in this study was the softmax function (Iglovikov et al., 2018). We use VGG16 backbone for all of segmentation models, it can greatly improve the performance of models (wang et al., 2023).

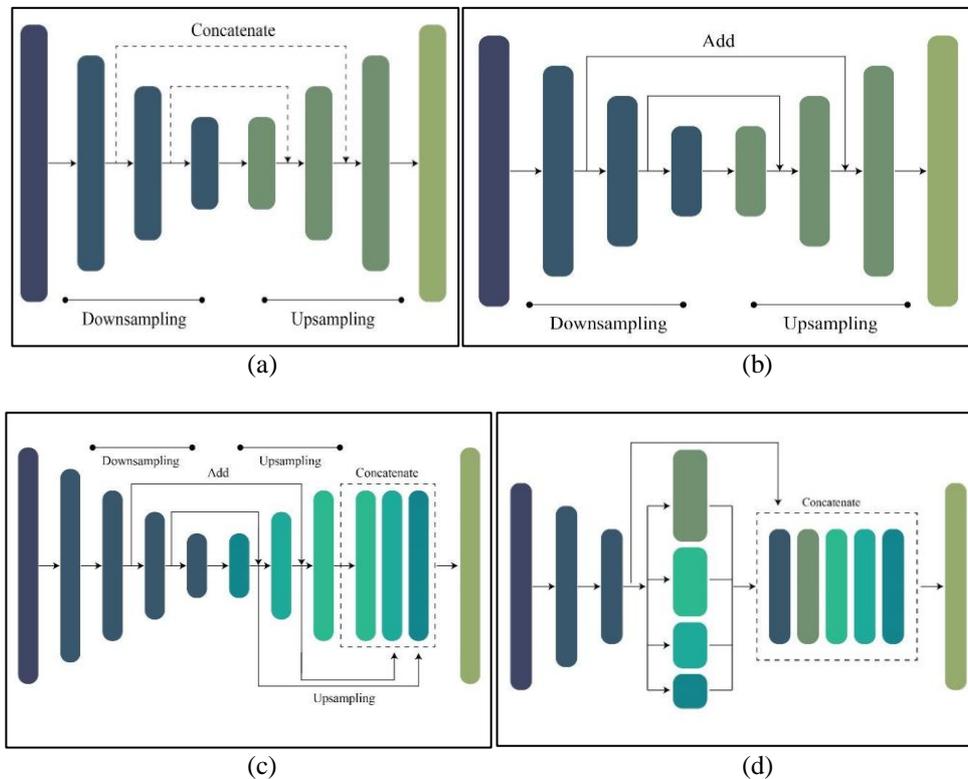

Figure 3. Segmentation models investigated in the study: (a) Unet (b) Linknet (c) FPN (d) PSPnet.

The UNet architecture is widely recognized as a key and influential model for semantic segmentation tasks. Initially, its intended application was in the field of biological imaging. However, its relevance has expanded significantly, and it is now being utilized in several domains of interest, such as remote sensing (Hu et al., 2020). The UNet architecture uses an encoder to extract features at several levels. Subsequently, the decoder combines these learned features and resolution through a complex stacking mechanism, which considers localization and feature representation (Ronneberger et al., 2015). The LinkNet architecture enhances the UNet framework by using upsampled feature representations containing resolution information instead of just concatenating them (Chaurasia et al., 2017). The remaining two Pyramid Networks endeavor to establish a hierarchical structure resembling a pyramid. PSPNet accomplishes this objective by constructing a pyramid structure by applying variable pooling on the lowest downsampled map (Zhao et al., 2017). This process generates diverse spatial resolutions, subsequently employed to enhance the features. In contrast, the FPN methodology operates by constructing two pyramids and later merging them to produce segmentation maps abundant in features at every level (Lin et al., 2017).

## 2.3 Evaluation Metrics:

The testing dataset, including 15% of the total dataset, was utilized to compute the algorithm output assessments, commonly called model evaluation. Three metrics were calculated to evaluate the model: intersection over union (IoU) (Rezatofighi et al., 2019), overall accuracy (OA), and F1-score (F1) (Sokolova et al., 2006). These metrics were calculated for three distinct classes such as other, vegetation, and water in Landsat and high spatial aerial image, while 9 classes such as asphalt, meadows, gravel, trees, metal sheets, bare soil, bitumen, bricks, and shadows in hyperspectral imagery. The IoU, or Intersection over Union, alternatively referred to as the Jaccard index, is extensively employed in semantic segmentation tasks (Rezatofighi et al., 2019). The Intersection over Union (IoU) is a metric used to measure the degree of overlap between visually interpreted labels and projected results. It is calculated by dividing the overlap area by the union area between ground truth and forecasted results. Table 6 displays the mathematical expressions for the model assessment metrics. In this context, the variables C, TP, TN, FP, and FN correspond to the class, true positive, true negative, false positive, and false negative, respectively.

$$\text{OA} = \frac{\sum_{i=1}^{C} TP_i}{\sum_{i=1}^{C}(TP_i + FP_i + TN_i + FN_i)} \quad (1)$$

$$\text{IoU} = \frac{|Ground\ truth\ \cap\ Predicted\ result|}{|Ground\ truth\ \cup\ Predicted\ result|} \quad (2)$$

$$F1 = 2\ x\ \frac{\sum_{i=1}^{C}\left(\left(\frac{TP_i}{TP_i+FP_i}\right)x\left(\frac{TP_i}{TP_i+FN_i}\right)\right)}{\sum_{i=1}^{C}\left(\left(\frac{TP_i}{TP_i+FP_i}\right)+\left(\frac{TP_i}{TP_i+FN_i}\right)\right)} \quad (3)$$

Where  $TP$ = true positif
$FP$ = false positif
$TN$ = true negatif
$TP$ = true positif

## 3. Result and Discussion

The models were trained using an early stopping schema to avoid overfitting. This indicates that the training process would stop automatically if there were no further improvements in the loss. Figure 4 compares IOU score and loss curves of each segmentation model at every epoch, and this points out that the loss reduces rapidly.

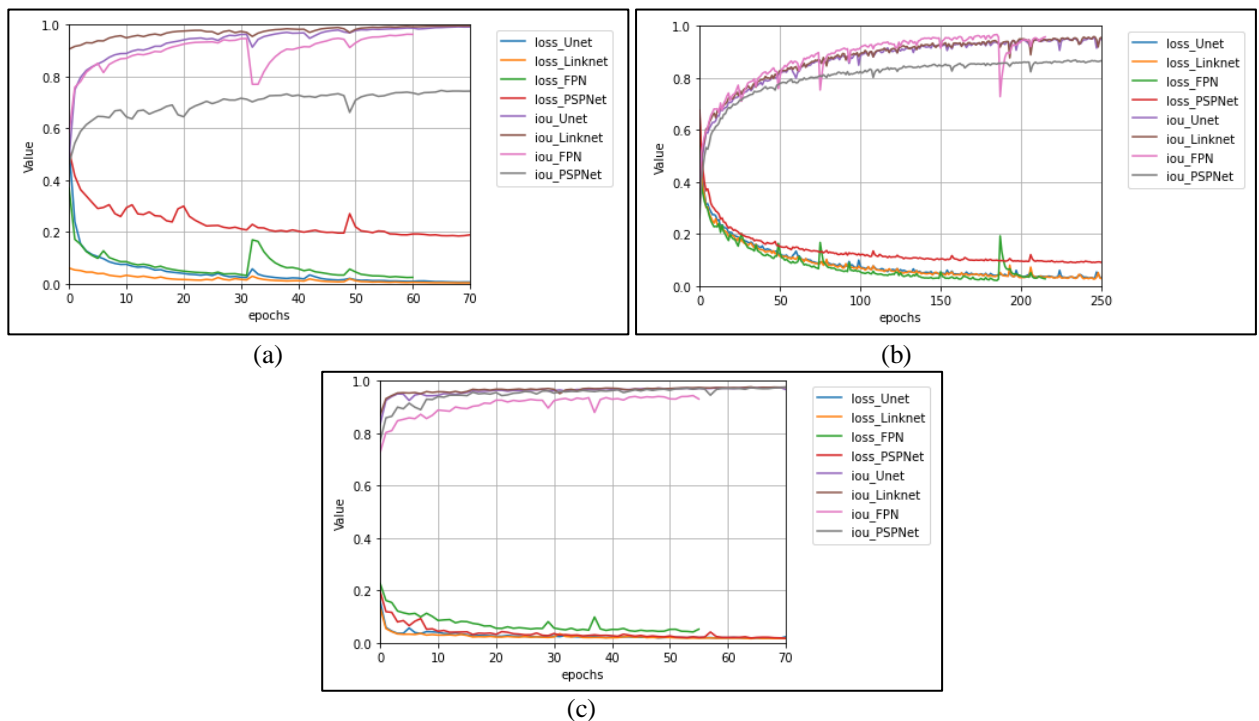

Figure 4. Loss and IoU Score, (a) Multispectral dataset, (b) Hyperspectral dataset and (c) High Spatial dataset.

The classification result from different segmentation models, namely U-Net, LinkNet, FPN, and PSPnet. We used the same input data and same seed random number to train the segmentation models. A visual comparison of a small part from the segmentation models result is presented in Figure 5, Figure 6, and Figure 7. Based on the visual analysis result, all of the four semantic segmentations achieved promising results on all three datasets. In the medium resolution satellite imagery (Landsat 8), the four semantic segmentation models can clearly distinguish 3 land cover classes (water, vegetation, other), while UNet and LinkNet have more clearly result based on visual analysis in Landsat 8 dataset. For the hyperspectral image result, the UNet, LinkNet, and FPN have similar visual result, but the PSPNet cannot clearly distinguish the detailed object. For the high resolution aerial imagery, all of the semantic segmentations have good visual analysis when we compared with the ground truth image.

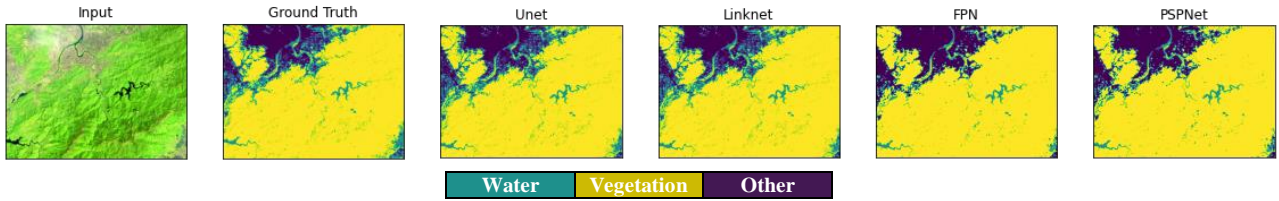

Fig. 5. Spatial distribution of training and testing sets, and the classification maps obtained by different models on the multispectral dataset.

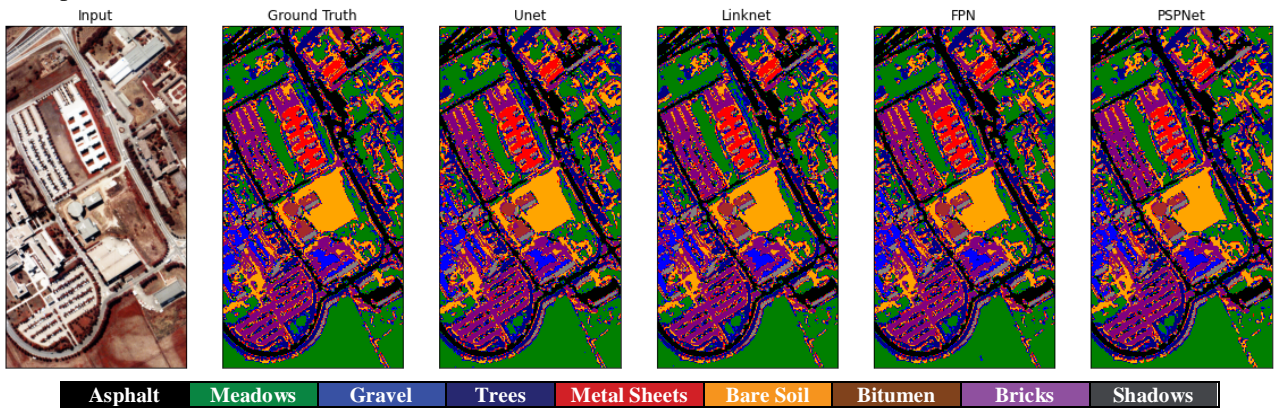

Fig. 6. Spatial distribution of training and testing sets, and the classification maps obtained by different models on the hyperspectral dataset.

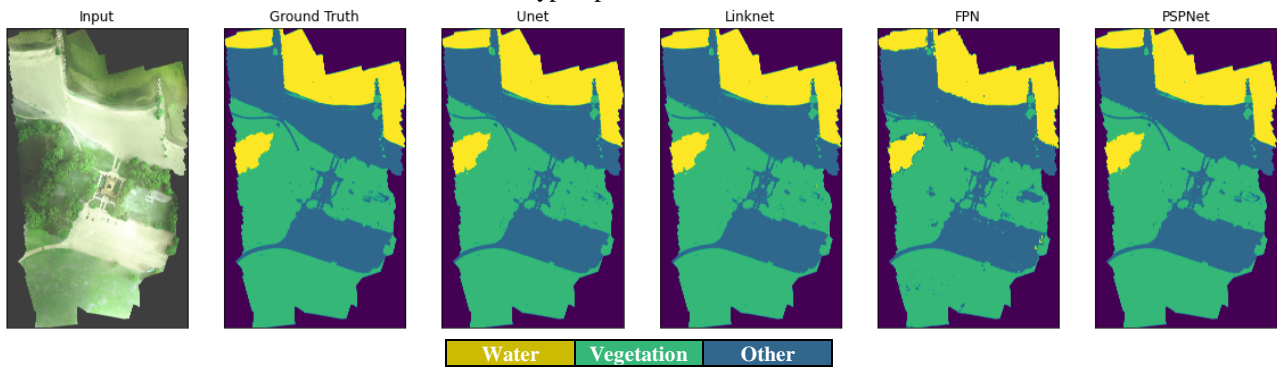

Fig. 7. Spatial distribution of training and testing sets, and the classification maps obtained by different models on the high spatial dataset.

The detailed quantitative comparison including Overall Accuracy, IoU score, and F1-score parameters as shown in Table 1. The best performing Model is found to be Linknet. When applying the Model for land cover, best performing imagery can be taken to be multispectral. The Linknet model outperforms other existing segmentation models in terms of Overall Accuracy, IoU Score, and F1-score are (multispectral: 0.99911, hyperspectral: 0.97264, high spatial: 0.99059), (multispectral: 0.99351, hyperspectral: 0.94162, high spatial: 0.98221), and (multispectral: 0.99733, hyperspectral: 0.96976, high spatial: 0.99099), respectively. The multispectral dataset outperforms other existing datasets in terms of Overall Accuracy, IoU Score, and F1-score are 0.99911, 0.99351, and 0.99733, respectively.

Table 1. Segmentation models exploration in terms of Overall Accuracy(OA), IoU Score, and F1-score.

| Dataset | Model | Overal Accuracy | IoU Score | F1-Score |
| --- | --- | --- | --- | --- |
| Multispectral | Unet | 0,99845 | 0,99164 | 0,99623 |
| | Linknet | **0,99911** | **0,99351** | **0,99733** |
| | FPN | 0,94655 | 0,71934 | 0,82120 |
| | PSPNet | 0,94585 | 0,71365 | 0,81522 |
| Hyperspectral | Unet | 0.97044 | 0.93515 | 0.96629 |
| | Linknet | **0.97264** | **0.94162** | **0.96976** |
| | FPN | 0.97164 | 0.93798 | 0.96783 |
| | PSPNet | 0.92271 | 0.83959 | 0.91201 |
| High spatial | Unet | 0.99050 | 0.98219 | **0.99103** |
| | Linknet | **0.99059** | **0.98221** | 0.99099 |
| | FPN | 0.95907 | 0.92602 | 0.96110 |
| | PSPNet | 0.99003 | 0.98130 | 0.99053 |

Our findings demonstrate that deep learning semantic segmentation achieves the promising accuracy for land cover mapping across different types of aerial imagery. We observed that these models consistently outperformed traditional methods, showcasing their ability to capture intricate spatial and spectral patterns in the data. In evaluating deep learning performance, the segmentation model architectures Linknet have better results in land cover segmentation using hyperspectral, multispectral, and high spatial aerial imagery. Due to segmentation accuracy of the network architecture when using in the hyperspectral and high spatial lower than multispectral dataset, hyperspectral and high spatial can be used to detect and extract more material.

## 4. Conclusion

It is possible to classify the land cover image using various deep learning methods. Unet, Linknet, FPN, and PSPNet, four deep learning segmentation model architectures frequently used in image classification, offer a classification model with an accuracy rate of more than 0.92. Linknet provides a model with an accuracy rate of more than 0.97, the highest of the four approaches, according to the comparison's findings. Furthermore, as remote sensing technology advances, future research should focus on developing more efficient and accurate deep learning models that especially use small labels for training to address the growing demands of land cover mapping and environmental monitoring applications.